\documentclass[letterpaper, 10 pt, conference]{ieeeconf}  

\IEEEoverridecommandlockouts                              

\usepackage{graphicx} 
\usepackage{mathtools}

\usepackage{amsmath} 

\usepackage{textcomp}
\usepackage{tabularx,booktabs}
\usepackage{color}
\usepackage[pdftex,dvipsnames]{xcolor}
\usepackage{array}
\usepackage{algorithm} 
\usepackage{algpseudocode} 
\usepackage{subcaption} 

\newcolumntype{P}[1]{>{\centering\arraybackslash}p{#1}}
\newcolumntype{C}{>{\centering\arraybackslash}X} 

\usepackage{tikz}

\newcommand*{\figtiTFlitefont}{\fontfamily{phv}\selectfont}

\title{\LARGE \bf
A-MuSIC: An Adaptive Ensemble System For Visual Place Recognition In Changing Environments
}

\author{Bruno Arcanjo$^{1}$, Bruno Ferrarini$^{1}$, Michael Milford$^{2}$, Klaus D. McDonald-Maier$^{1}$ and Shoaib Ehsan$^{1, 3}$
\thanks{}
\thanks{$^{1}$B. Arcanjo, B. Ferrarini, K. D. McDonald-Maier and S. Ehsan are with the School of Computer Science and Electronic Engineering, University of Essex, United Kingdom {\tt\small (email: bq17319@essex.ac.uk; bferra@essex.ac.uk; kdm@essex.ac.uk; sehsan@essex.ac.uk)}}%
\thanks{$^{2}$M. Milford is with the School of Electrical Engineering and Computer Science, Queensland University of Technology, Brisbane, QLD 4000, Australia
        {\tt\small (email: michael.milford@qut.edu.au)}}%
\thanks{$^{3}$S. Ehsan is also with the school of Electronics and Computer Science, University of Southampton, United Kingdom \tt\small{(email: s.ehsan@soton.ac.uk)}}%
}

\begin{document}

\maketitle
\thispagestyle{empty}
\pagestyle{empty}

\begin{abstract}
Visual place recognition (VPR) is an essential component of robot navigation and localization systems that allows them to identify a place using only image data. VPR is challenging due to the significant changes in a place’s appearance under different illumination throughout the day, with seasonal weather and when observed from different viewpoints. Currently, no single VPR technique excels in every environmental condition, each exhibiting unique benefits and shortcomings. As a result, VPR systems combining multiple techniques achieve more reliable VPR performance in changing environments, at the cost of higher computational loads. Addressing this shortcoming, we propose an adaptive VPR system dubbed Adaptive Multi-Self Identification and Correction (A-MuSIC). We start by developing a method to collect information of the runtime performance of a VPR technique by analysing the frame-to-frame continuity of matched queries. We then demonstrate how to operate the method on a static ensemble of techniques, generating data on which techniques are contributing the most for the current environment. A-MuSIC uses the collected information to both select a minimal subset of techniques and to decide when a re-selection is required during navigation. A-MuSIC matches or beats state-of-the-art VPR performance across all tested benchmark datasets while maintaining its computational load on par with individual techniques.
\end{abstract}

\section{Introduction}
\label{intro}

\begin{figure}[thpb]
\vspace*{1ex}
\centering
\textsc{\small\figtiTFlitefont{A-MuSIC System}}\par\vspace*{3ex}
\includegraphics[width=1.0\columnwidth]{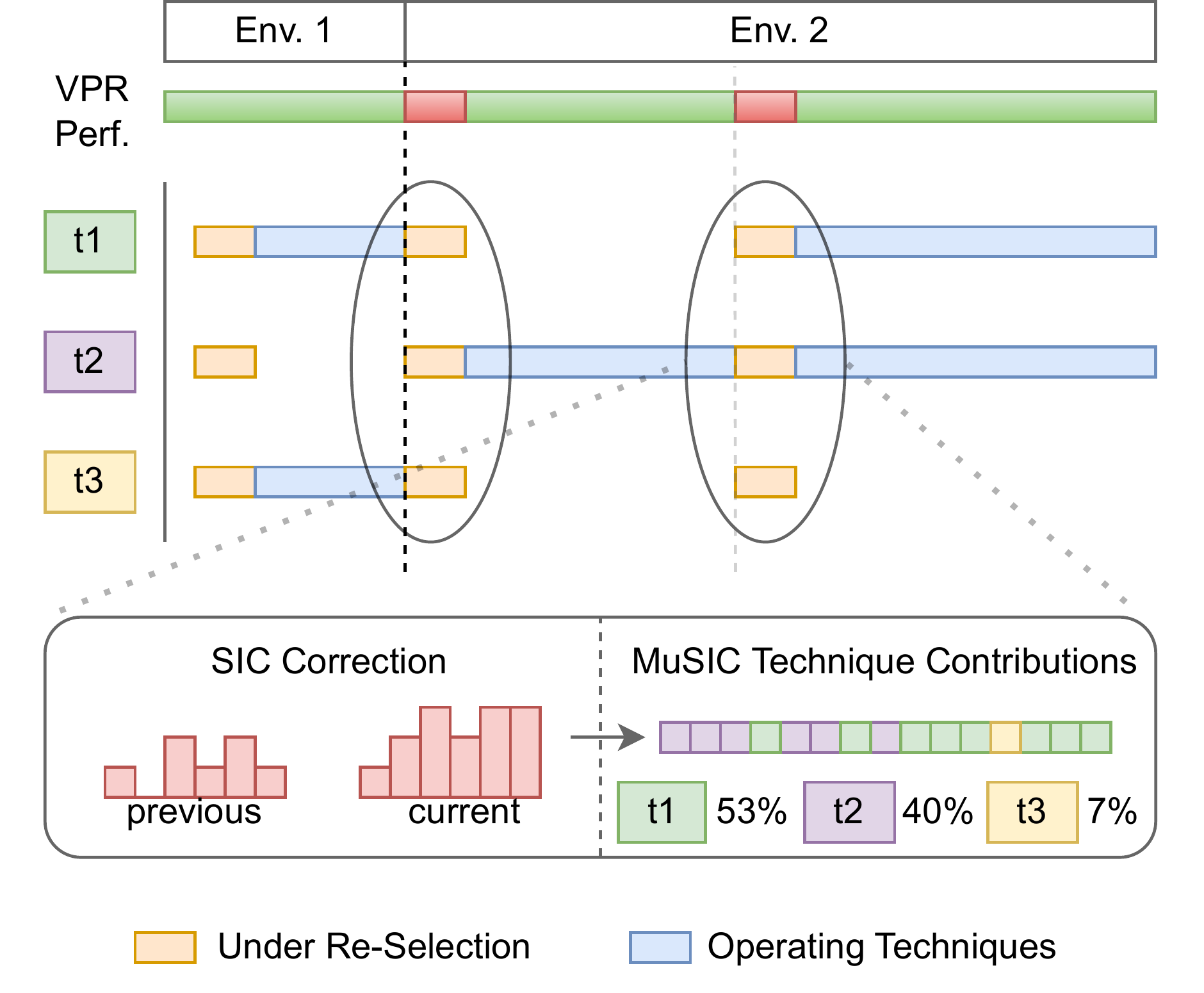}
\caption{A-MuSIC uses correction information obtained by SIC to detect a detriment in VPR performance, triggering a re-selection. Only the minimal amount of techniques needed for the current viewing conditions is selected, saving computational resources.}
\label{fig:visual_abstract}
\end{figure}

Visual place recognition (VPR) aims to solve the localization component of Simultaneous Localization and Mapping (SLAM) using image information, attractive feature due to the low cost, versatility and availability of cameras \cite{ref:vpr-survey}. However, VPR is challenging due to the variety of ways in which a place's appearance can change. Changes in illumination \cite{ref:illu_changes}, seasonal variations \cite{ref:season_changes}, and varying viewpoints \cite{ref:pov_changes} can make the same place appear vastly different. Many techniques have been proposed to tackle these challenges, but no standalone technique excels in every viewing condition \cite{ref:maria_compl}.

Combining multiple techniques into a single VPR algorithm can compensate for individual weaknesses, as demonstrated by systems such as \cite{ref:mpf, hier_mpf, switchhit}. However, these systems either run a static number of techniques for every query frame \cite{ref:mpf, hier_mpf}, potentially wasting computational resources, or rely on extra ground-truth information of the runtime environment \cite{switchhit}.

In this work, we address the shortcomings of existing combination methods by proposing an adaptive VPR system that selects the minimal optimal subset of techniques based on their runtime VPR performance, without extra ground-truth information. The core of the proposed approach consists of selecting the most competent techniques for the current viewing conditions and triggering a re-selection when those conditions change enough to degrade the VPR capabilities of the running subset. We develop Self-Identification and Correction (SIC), an algorithm for assessing the runtime performance of a technique without extra ground-truth information. SIC identifies when a technique has wrongly matched a query frame, proposes an alternative prediction and quantifies this correction. SIC leverages the frame-to-frame similarity continuity that a correct match should present in sequential navigation to filter out erroneous prediction candidates proposed by a VPR technique.

SIC can operate with multiple VPR techniques, a process dubbed Multi-SIC (MuSIC). MuSIC runs and corrects all VPR techniques individually, while generating their respective correction information, and then selects which technique to trust on a per-frame basis. This per-frame selection information is used to identify which techniques are significantly contributing to the current navigation.

Finally, we present our adaptive VPR system, dubbed Adaptive Multi-SIC (A-MuSIC), whose functionality is visible in Fig. \ref{fig:visual_abstract}. A-MuSIC continuously analyses the SIC correction being performed by the active technique subset. A statistically significant change in correction signifies a change in the VPR performance of the active subset, triggering a re-selection. During the re-selection stage, all techniques in the ensemble are evaluated and only the most selected ones by MuSIC remain active until the next re-selection stage.

A-MuSIC effectively saves computational resources by only running the minimally required subset of techniques for the current viewing conditions, while still enjoying the VPR performance benefits of multi-technique systems. It is capable of operating with as little as a single technique up to the entire ensemble, requiring no extra ground-truth information of the runtime environment.

Accordingly, we claim the following contributions:

\begin{itemize}
    \item A novel algorithm, SIC, which identifies incorrect VPR technique matches, proposes new predictions and assesses online technique performance.

    \item MuSIC, an extension of SIC that leverages a static ensemble of VPR techniques to determine the most suitable match for a given query frame, quantifying technique contributions.
    
    \item A-MuSIC, an adaptive VPR system that optimizes the selection of VPR techniques during runtime to ensure that only the currently necessary techniques are active, saving computational resources while maintaining performance benefits.
\end{itemize}

The rest of this letter is organized as follows. In Section \ref{relatedwork} we provide an overview of different approaches to VPR, with a focus on technique combination and sequence-based methods. Section \ref{method} describes our methodology, starting from the online check and correction of a single technique, followed by the extension to a static number of techniques, and finally the implementation of the adaptive selection algorithm. In Section \ref{exp_setup} we detail the settings of our experimentation. We show and analyse our results in Section \ref{results}. In Section \ref{conclusions}, we highlight the benefits and limitations of our method and suggest future research paths.

\section{Related Work}
\label{relatedwork}
Appearance-based localization continues being an important research topic, with several approaches being proposed in the literature. The underlying technology on which these techniques are based on varies widely. 

\cite{ref:fabmap} utilizes hand-crafted feature descriptors, such as Scale Invariant Feature Transform \cite{ref:sift} and Speeded Up Robust Features \cite{ref:surf}, to successfully build a landmark representation of the environment. While the use of local features improves resilience against viewpoint variations, it makes the method sensitive to appearance changes such as different illumination conditions \cite{cummins2008fabmap}. Well-studied global descriptors  like Histogram of Oriented gradients \cite{ref:hog} and \cite{ref:gist} have also been employed in VPR \cite{ref:mcmanus2014scene} but struggle with viewpoint changes. CoHOG \cite{ref:cohog} utilizes a region-of-interest approach in conjunction with HOG to minimize this shortcoming. \cite{rmac} is another popular VPR technique based on the computation of regions of interest in an image.

Image features retrieved from the inner layers of Convolutional Neural Networks (CNNs) have been shown to outperform hand-crafted features \cite{ref:cnn_for_vpr1, ref:cnn_for_vpr2}. Techniques such as HybridNet, AMOSNet \cite{ref:hybridasmosnet} and NetVLAD \cite{ref:netvlad} have utilized these CNN-extracted features to perform state-of-the-art VPR. Computational efficiency is another added consideration, with CALC \cite{ref:calc} being an example of a CNN-based VPR technique designed to perform lightweight VPR.

The variety in available image-processing methods, with different sets of strengths and weaknesses, has led to recent research on the combination of multiple techniques to perform VPR in changing environments. Multi-process fusion (MPF) \cite{ref:mpf} proposes a system which combines four VPR methods utilizing a Hidden Markov Model to further infuse sequential information. \cite{hier_mpf} combines techniques in a hierarchical structure, passing only the top place candidates of upper level techniques down to the lower tiers. SwitchHit \cite{switchhit} instead starts by running a single technique and, using prior knowledge of VPR performance in the environment, decides if another technique should be run, repeating the process until a satisfactory confidence threshold is achieved.

In this letter, we propose a multi-technique VPR system which is able to dynamically select which VPR techniques should be used in the current environmental conditions. Unlike \cite{ref:mpf, hier_mpf}, our system can operate with as low as a single technique if the remaining are not contributing to the VPR task. Moreover, it does not require any additional ground-truth information about the deployment environment nor technique complementarity, a major benefit over \cite{switchhit}.

\section{Methodology}
\label{method}

Our proposed adaptive system A-MuSIC relies on (a) identifying the online performance of VPR techniques and (b) determining which techniques are necessary for reliable VPR in the current environment. Point (a) is directly addressed by SIC, which corrects a VPR technique by analysing the similarity continuity of its past queries and quantifies the performed correction. Point (b) is then tackled by MuSIC, which selects the most trustworthy technique on a per-frame basis and is thus able to quantify the contribution of individual techniques. In this section, we detail the operation of SIC, MuSIC and how A-MuSIC makes use of the collected information to perform adaptive VPR.

\subsection{Self-Identification and Correction (SIC)}
\label{self_identifying_self_correcting}

\begin{figure}[t]
\vspace*{1ex}
\centering
\textsc{\small\figtiTFlitefont{SIC With K=2, F=2}}\par\vspace*{3ex}
\includegraphics[width=0.8\columnwidth]{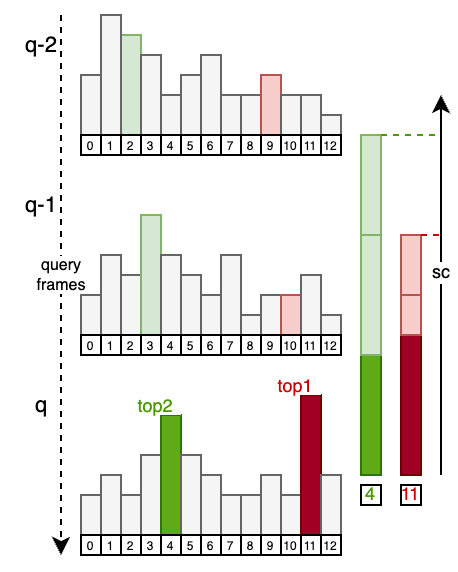}
\caption{The top match for frame $q$ (solid red) suggests a wrong match and SIC corrects it to the second highest score (solid green).}
\label{fig:sic_fig}
\end{figure}

The ability to identify a mismatched place at runtime remains an open but important problem in VPR, as it allows for assessing the online performance of a running VPR technique. Moreover, detecting wrongly matched frames indicates that the query should be re-matched, directly increasing VPR performance given a successful correction. In real-world applications, ground-truth information is of the deployment environment is rarely available, hence extra information and assumptions are utilized to estimate when an online match is incorrect. Our proposed SIC method quantifies the sequential consistency of the place matching similarity distributions of a technique over a number of past query frames, for both identifying incorrect matches and attempting their correction. 

During navigation, when VPR is performed for a query frame ${q}$, the employed technique generates a score distribution ${S_q}$ where each score ${S_{q, i}}$ corresponds to one of the ${N}$ reference places. With descriptor-based methods, such as HybridNet or NetVLAD, these scores are often the cosine differences between the query frame and the reference images. The usual approach is to take the reference place associated with the highest score in ${S_q}$, i.e. ${argmax(S_q)}$, as the place prediction for the evaluated frame \cite{ref:vpr_uavs}. However, we empirically observe that, even in incorrect matches, the score of the correct reference place is often relatively high. As the correct match is often found among the top scores, we interpret the ${top_K}$ values as possible match candidates for the current frame, possibly misplaced due to visual noise. We evaluate the sequential consistency $sc$ of each of these top-scoring candidates by referring to the similarity distributions of previous queries. This concept is illustrated in Fig. \ref{fig:sic_fig}, where a match is successfully corrected by computing the $sc$ of the $top_1$ and $top_2$ scores. 

We now detail SIC for the correction of a query frame ${q}$. ${S}$ is a matrix where each row vector ${S_q}$ is a score distribution, one per frame for which VPR was performed during the current navigation. Importantly, these rows are ordered by time of observation, hence ${S_{q, i}}$ represents the ${i^{th}}$ score of the ${q^{th}}$ performed query. The ${top_K}$ scores are selected for analysis from ${S_{q}}$, where ${K}$ is a hyperparameter. ${F}$, also an hyperparameter, denotes the maximum number of past queries we want to consider when computing the $sc$ of each top candidate. For each of these candidates $c$, a lower bound ${f_{lb}}$ is set to prevent trying to look at observation indices that do not exist, i.e. before the start of navigation. Starting from the observation vector ${S_{p}}$, the algorithm shifts $\delta$ observations in $S$ to $S_{p+\delta}$, for all valid $\delta$ values. Note that $\delta$ takes on negative values, representing a shift to previously performed queries stored in $S$. With each query shift, SIC also shifts $\delta$ reference place indices from the current top candidate ${c}$ within the vector $S_{p+\delta}$ (Fig. \ref{fig:sic_fig}). The score $S_{q+\delta, c+\delta}$ is added to the rolling sum $sc$. Finally, we take the candidate which achieved the highest average $sc$ as the final place prediction. If this candidate is not the original top ranked score, we consider the original matching an error which was corrected.

Assuming sequential navigation, SIC looks at $\delta$ query results back from the current frame. The position of the related score peaks are expected to appear back-shifted accordingly by $\delta$ indices. By repeating the shift, summing the score at the shifted indices, we quantify the candidate's frame-to-frame continuity, i.e. its $sc$. The difference between the average $sc$ of the erroneous match and the corrected prediction can then be collected as the quantification of the correction for assessing online performance. Moreover, the correction process itself can improve VPR performance as more places are matched correctly, as observed in Fig. \ref{fig:std_vs_sic_accs}.

\subsection{Multi-Technique SIC (MuSIC)}
\label{MuSIC_expl}
The motivation for utilizing multiple VPR techniques in unison is increasing tolerance against various types of visual changes. Specifically in Multi-SIC, techniques that perform well within the current environment achieve higher $sc$ than techniques that perform poorly. Hence, MuSIC operates by comparing the $sc$ values of each technique's corrected prediction. Furthermore, there is the added benefit of a higher chance that the correct match is within the $top_K$ candidates of at least one of the techniques.

Different techniques have different ranges of output similarity scores. Since $sc$ is computed directly from these scores, we scale the score vectors to allow for a fair comparison. Therefore, each score ${S_{q, i}}$ is normalized to the same value range using the following equation:
\begin{gather}
  {\hat{S}}_{q, i} = \frac{S_{q, i} - \mu}{\sigma}
\end{gather}
\noindent where ${\hat{S}}_{q, i}$ is the output scaled value, $\mu$ and $\sigma$ are the mean and standard deviation of vector ${S_q}$ currently being scaled, respectively. Note that this operation does not affect the behaviour of SIC but it is crucial to compare $sc$ values between different techniques.

Each technique then independently corrects itself and proposes its top candidate with highest $sc$. The technique whose candidate achieves the highest $sc$ is chosen and its proposed prediction is trusted to be the correct prediction for the given query frame. Since MuSIC only chooses one technique per frame, keeping track of its choices during navigation allows for quantifying the contributions of each individual technique.

\subsection{Adaptive MuSIC (A-MuSIC)}
\label{AMUSIC}
By using a static ensemble of techniques, MuSIC achieves higher VPR performance across different environments (Fig. \ref{fig:pr_curves}. However, using all techniques for every frame is computationally expensive and potentially wasteful if a smaller subset would have sufficed for the current visual conditions. Our adaptive system A-MuSIC addresses this drawback by dynamically selecting the optimal subset of techniques to be run in the current environment. The next subsections detail the selection process and reselection mechanism upon detriment of VPR performance.

\subsubsection{Technique Selection}
\label{selection}

As explained in \ref{MuSIC_expl}, MuSIC performs a post-correction selection of the place prediction of a set, $T$, of VPR techniques. MuSIC tracks the history, $th$, of the selected techniques over the past $M$ frames. Every element in $th$ corresponds to a VPR technique in $T$. From ${th}$, we calculate the proportion of selection of each technique, dubbed $coverage$ and immediately add the technique with highest $coverage$ to the subset of active techniques $\hat{T}$. We repeatedly add techniques to $\hat{T}$ by order of $coverage$, until it cumulatively reaches a pre-designated threshold, $E$. Higher $E$ values favour VPR performance, usually at increased computational cost, as the system is more likely to select multiple techniques. Note that it is possible to choose as little as one technique even for a high $E$ value, given the case that a single technique achieves high enough $coverage$. Conversely, it is also possible that all techniques are selected, depending on the $coverage$ distribution per technique.

\subsubsection{Re-Selection Trigger}
\label{re-selection}

In the first $M$ frames of navigation, all techniques in $T$ are utilized and SIC generates a correction history vector $ch$ of size $M$ per technique. Each element $ch_m$ in $ch$ is the per technique $sc$ difference between the corrected prediction for frame $m$ and its uncorrected prediction. Therefore, $ch_m$ is $0$ when no correction was performed, with larger values indicating a larger discrepancy between the corrected and uncorrected prediction. A selection is then performed, constructing the first subset $\hat{T}$ and the $ch$ vectors of the selected techniques are concatenated into a single vector.

During navigation, every $M$ frames, the techniques in $\hat{T}$ each generate a new correction history vector $\hat{ch}$ which are again concatenated. The goal is to identify when the previous correction information stored in $ch$ is significantly different from the current correction, stored in $\hat{ch}$. We use a paired-sample, two-tailed T-test between the two distributions, at a significance level of $5\%$. The null hypothesis $H_0$ is that the population means of $ch$ and $\hat{ch}$ are equal and the alternative hypothesis $H_1$ is that they are significantly different. Importantly, the same technique subset $\hat{T}$ computes both $ch$ and $\hat{ch}$. Therefore, rejecting the null hypothesis represents a significant change in the correction behaviour of techniques in $\hat{T}$, which is interpreted as a change in the VPR performance of the subset. When $H_0$ is rejected, a new selection process is triggered using the same $M$ frames that indicated the change of environment, resulting in a possibly new $\hat{T}$. When a re-selection is not triggered, $ch$ is simply updated to equal $\hat{ch}$. If a re-selection does take place, $ch$ is updated to equal the correction vectors of the newly selected techniques. 

\section{Experimental Setup}
\label{exp_setup}
We conduct experiments to evaluate single technique SIC, MuSIC and the adaptive system A-MuSIC, analysing how each increment of the system affects VPR performance and computational demand. The rest of this section gives details on datasets, baseline VPR techniques utilized, hyperparameter ablation studies and evaluation metrics.

\subsection{Datasets}
\label{datasets}

We evaluate our approach on five different benchmark datasets: Nordland Winter and Fall \cite{ref:nord}, Gardens Point, St. Lucia \cite{stlucia}, 17 Places \cite{17places} and Berlin \cite{berlin}. Table \ref{datasets} provides details on how we utilize these datasets. We use a ground-truth tolerance of 1 frame for all datasets except for 17 Places where we use a tolerance of 10 frames.

\newcolumntype{M}[1]{>{\centering\arraybackslash}m{#1}}
\begin{table}[t]
  \centering
  \vspace{5px}
  \caption{Dataset Details}
  \label{tab:datasets}%
    \begin{tabular}{M{1cm}M{1.5cm}M{1.5cm}M{1.5cm}M{1cm}}
    \toprule
Dataset  & Condition   & Reference Traverse  & Query Traverse & Number of Images \\
\midrule
\midrule
Nordland Winter  & Extreme seasonal    & Summer   & Winter      &   1000      \\
\midrule
Nordland Fall        & Moderate seasonal   & Summer &  Fall       &  1000   \\
\midrule
Berlin     &  Strong viewpoint & halen.-2, kudamm-1 and A100-1   &  halen.-1, 
kudamm-2 and A100-2 & 250  \\
\midrule
Night-Right & Outdoor Illumination; Lateral Shift & Day-Left     & Night-Right     &  200 \\
\midrule
St. Lucia   & Daylight; Dynamic Elements &   Afternoon   &    Morning    &   1100  \\
\midrule
17 Places  & Indoor Illumination & Day     &  Night      &    2000  \\

\bottomrule
\end{tabular}
\end{table}%

\subsection{Evaluation}
\label{eval_metrics}

\subsubsection{Precision-Recall}
VPR performance is often quantified using Precision-Recall (PR) curves and the area under these curves (AUC) \cite{ref:pr_jus1}. The use of PR curves and respective AUC is favoured for class imbalanced datasets. In the VPR context, a small set of correct predictions and a much larger set of incorrect predictions exists for each query frame, resulting in a strongly imbalanced dataset.

\subsubsection{Accuracy}
When evaluating SIC, we are more interested in its match correction capabilities rather than practical VPR performance. While correcting wrong matches leads to improved VPR performance captured by AUC, accuracy is more suitable to analyse the percentage of SIC corrected frames.

\subsubsection{Quantifying Adaptability}
\label{ptr_expl}
The main advantage of an adaptive VPR system is achieving tolerance against various types of visual changes while not relying on the brute-force usage of multiple baseline techniques. To assess the VPR performance benefits of A-MuSIC in changing environments, we additionally compare the average VPR performance across datasets of the tested techniques. Furthermore, we compute the proportion of technique runs, $PTR$, for A-MuSIC, given by

\begin{gather}
  {PTR} = \frac{1}{Q * |T|} \sum_{q=1}^{Q} |\hat{T_q}|
\end{gather}

\noindent where $Q$ is the total number of queries, $|T|$ is the cardinality of the starting set of techniques $T$ and $|\hat{T_q}|$ is the cardinality of subset of techniques $\hat{T}$ for query $q$. The maximum $PTR$ value is $1$ when all techniques in $T$ are used for every query frame. The minimum $PTR$ value is given by

\begin{gather}
  {Min PTR} = \frac{1}{|T|}
\end{gather}

\noindent which occurs when only one technique is employed for every query frame. With our configuration of four VPR techniques, the minimum possible $PTR$ is $0.25$. 

We employ $PTR$ to analyse the computational benefits of A-MuSIC and how the changes in total techniques usage affects VPR performance and prediction computation time.

\subsection{System Configuration}
\label{sys_config}

\subsubsection{Baseline Techniques}
A-MuSIC is not bound to any specific ensemble of VPR techniques, with any combination of any size being possible. In our experiments, we employ a starting ensemble of four VPR techniques: HOG, NetVLAD, CoHOG and CALC and we use the default implementations provided in \cite{ref:vpr_bench}. We note that neither CALC nor NetVLAD were directly trained on the employed benchmark datasets.

\subsubsection{A-MuSIC Settings}

\begin{figure}[!t]
\vspace*{1ex}
\centering
\includegraphics[width=0.8\columnwidth]{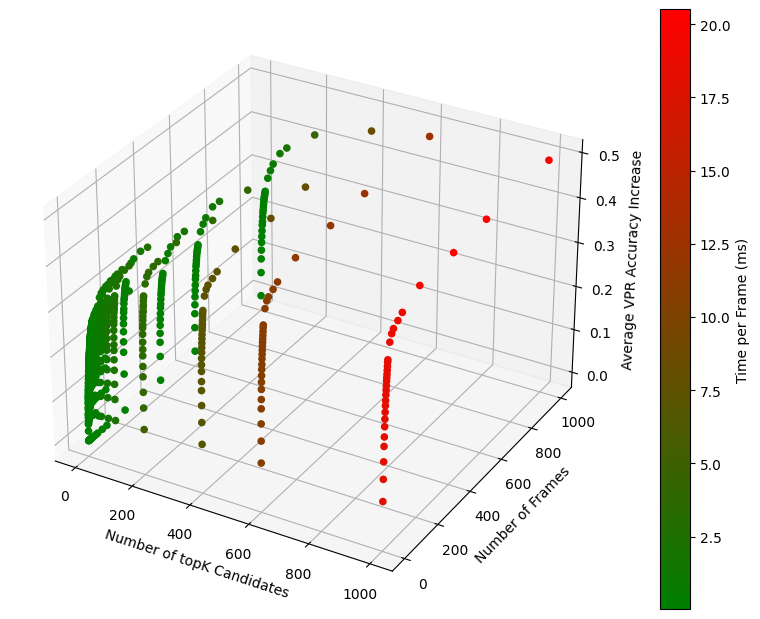}
\caption{Ablation study on the hyperparameters of SIC, $top_K$, $F$. and their effects on average VPR accuracy and prediction time per frame.}
\label{fig:sic_ablation}
\end{figure}

As detailed in Section \ref{method}, SIC contains several hyperparameters and we conduct ablation studies to find the best values in terms of VPR performance and computational efficiency. Rather than fine-tuning SIC for each technique and benchmark dataset, we select global parameter values that result in the best average VPR performance and computation trade-off. The results of the ablation study are presented in Fig. \ref{fig:sic_ablation}. We observe that most of the performance increase can be obtained with low $K$ values, with heavily diminishing returns for higher settings. Furthermore, due to the vectorization of the implementation, larger $F$ settings do not have a detrimental impact on the computational cost of SIC. Overhead computation time per correction is mostly affected by the number of $top_{K}$ candidates, ranging from less than 1 to 21 milliseconds. Since SIC is performed on top of multiple techniques and most of the performance benefit is seen on lower settings, we choose a relatively conservative $top_{K}$ value of $50$. As $F$ does not significantly affect runtime speed, we set its maximum value to $1000$ past observations frames.

As detailed in Section \ref{AMUSIC} the hyperparameters $E$ and $M$ are introduced with the implementation of A-MuSIC. The parameter $E$ controls the target $coverage$ threshold to achieve when selecting a subset of techniques. We set $E$ to $0.7$, signalling we require that at least $70\%$ of the selection frames to be covered by the subset of techniques. The setting $M$ defines how many selection frames are used and we select a value of $10$.

\section{Results \& Discussion}
\label{results}
In this Section, we present the results obtained by the intermediary components SIC and MuSIC as well as the fully-pledged A-MuSIC system.

\begin{table*}[htbp]
  \centering
  \vspace{5px}
  \caption{VPR Performance (AUC) and Prediction Time (ms)}
    \begin{tabular}{|c|cc|cc|cc|cc|cc|cc|cc|}
    \hline
          & \multicolumn{2}{c|}{\textbf{Winter}} & \multicolumn{2}{c|}{\textbf{Fall}} & \multicolumn{2}{c|}{\textbf{Berlin}} & \multicolumn{2}{c|}{\textbf{Night-Right}} & \multicolumn{2}{c|}{\textbf{17 Places}} & \multicolumn{2}{c|}{\textbf{St. Lucia}} & \multicolumn{2}{c|}{\textbf{Average}} \\
          \cline{2-15}
          & \multicolumn{1}{c|}{\textbf{AUC}} & \multicolumn{1}{c|}{\textbf{ms}} & \multicolumn{1}{c|}{\textbf{AUC}} & \multicolumn{1}{c|}{\textbf{ms}} & \multicolumn{1}{c|}{\textbf{AUC}} & \multicolumn{1}{c|}{\textbf{ms}} & \multicolumn{1}{c|}{\textbf{AUC}} & \multicolumn{1}{c|}{\textbf{ms}} & \multicolumn{1}{c|}{\textbf{AUC}} & \multicolumn{1}{c|}{\textbf{ms}} & \multicolumn{1}{c|}{\textbf{AUC}} & \multicolumn{1}{c|}{\textbf{ms}} & \multicolumn{1}{c|}{\textbf{AUC}} & \textbf{ms} \\
    \hline
    \hline
    HOG   & 0.29  & 49    & 0.84  & 48    & 0.03  & 16    & 0.03  & 15    & 0.34  & 96    & 0.56  & 47    & 0.35  & 45 \\
    CALC  & 0.30  & 168   & 0.88  & 168   & 0.06  & 157   & 0.13  & 161   & 0.38  & 170   & 0.48  & 165   & 0.37  & 165 \\
    CoHOG & 0.23  & 795   & 0.85  & 785   & 0.28  & 225   & 0.45  & 185   & 0.31  & 1525  & 0.37  & 729   & 0.42  & 707 \\
    NetVLAD & 0.28  & 723   & 0.68  & 719   & 0.81  & 722   & 0.54  & 717   & 0.48  & 727   & 0.35  & 721   & 0.52  & 722 \\
    HOG+SIC & 0.76  & 54    & 0.98  & 55    & 0.02  & 20    & 0.40  & 17    & 0.46  & 104   & 0.84  & 52    & 0.58  & 50 \\
    CALC+SIC & 0.86  & 172   & 0.99  & 175   & 0.56  & 163   & 0.71  & 163   & 0.67  & 177   & 0.92  & 170   & 0.79  & 170 \\
    CoHOG+SIC & 0.67  & 747   & 0.99  & 776   & 0.80  & 218   & 0.87  & 184   & 0.76  & 1518  & 0.84  & 718   & 0.82  & 694 \\
    NetVLAD+SIC & 0.78  & 747   & 0.94  & 743   & 0.96  & 736   & 0.97  & 742   & 0.82  & 745   & 0.85  & 717   & 0.89  & 738 \\
    MuSIC & 0.95  & 1790  & 1.00  & 1748  & 0.95  & 1164  & 0.97  & 1094  & 0.86  & 2544  & 0.92  & 1937  & 0.94  & 1713 \\
    A-MuSIC & 0.90  & 621   & 0.97  & 649   & 0.95  & 558   & 0.98  & 766   & 0.85  & 982   & 0.92  & 747   & 0.93  & 721 \\
    \hline
    \end{tabular}%
  \label{tab:vpr_perf_comp_times}%
\end{table*}%

\subsection{SIC}

\begin{figure}[!t]
\vspace*{1ex}
\centering
\includegraphics[width=1.0\columnwidth]{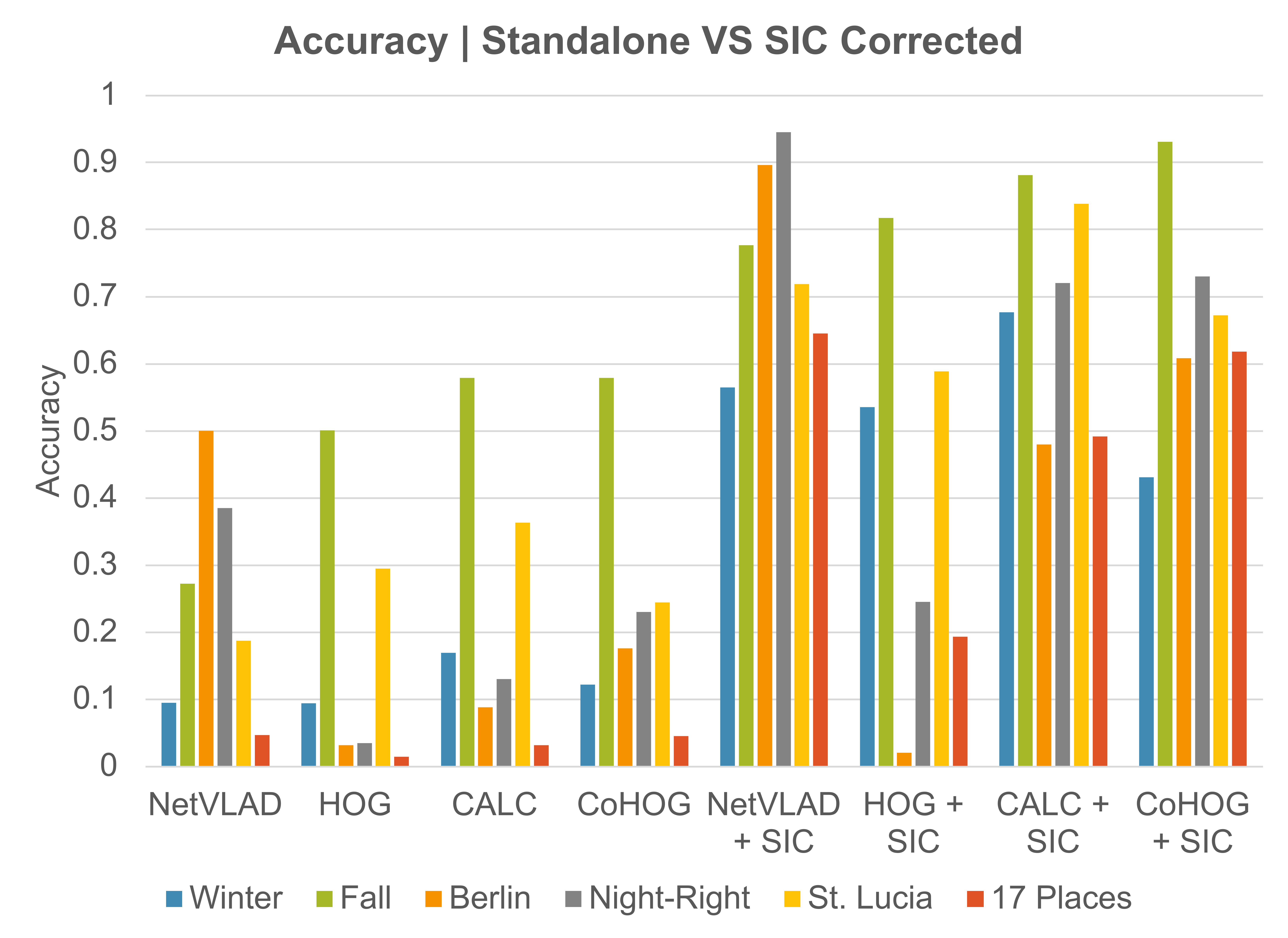}
\caption{VPR accuracies of individual VPR techniques and their SIC corrected counterparts.}
\label{fig:std_vs_sic_accs}
\end{figure}

In Fig. \ref{fig:std_vs_sic_accs}, we compare the VPR accuracies of the standalone VPR techniques against their SIC corrected counterparts. Nearly all techniques achieve higher VPR accuracy across all datasets, showing how SIC is able to successfully correct a large portion of wrong matches. The only exception is HOG on the Berlin dataset, where no accuracy improvement was achieved. This behaviour is likely due to how poorly HOG performs on the dataset, leaving no useful frame-to-frame continuity information for SIC to exploit. The accuracy increase, i.e. more correct matches, also results in increased VPR performance, which can be observed in Table \ref{tab:vpr_perf_comp_times}, with higher AUC values across the board.

Also in Table \ref{tab:vpr_perf_comp_times}, we observe the impact of SIC on the prediction time of the individual techniques. For techniques with high prediction times, such as NetVLAD or CoHOG in larger datasets, the addition of SIC is negligible, being within variation. In the case of faster techniques, such as HOG and CALC, a small average increase of 5 milliseconds is reported. 

\begin{figure*}[!t]
	\centering
	\begin{subfigure}[b]{0.32\textwidth}
		\centering
		\includegraphics[width=\linewidth]{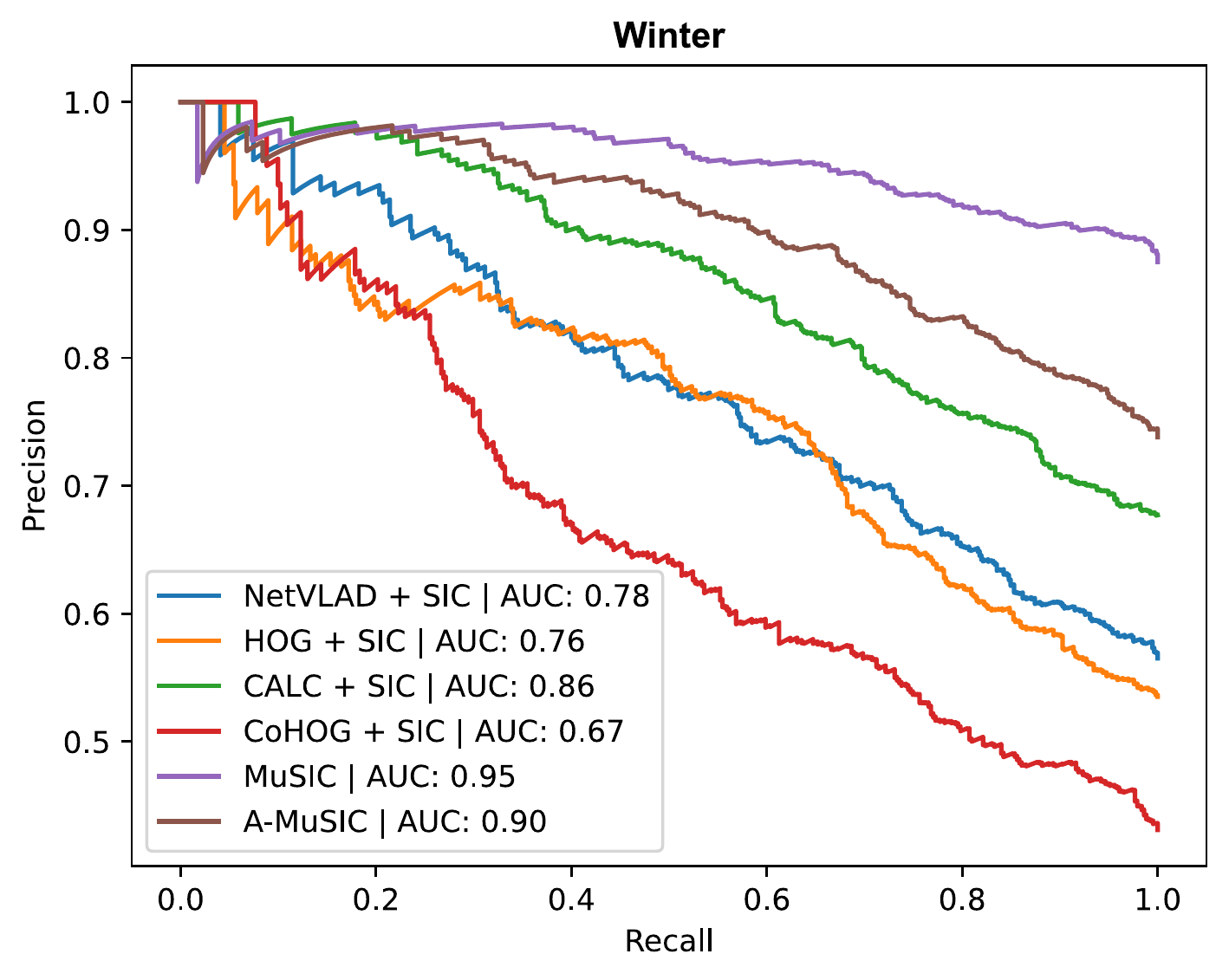}
		\caption{}
		\label{fig:pr_curves:A}
	\end{subfigure}
	\hfill
	\begin{subfigure}[b]{0.32\textwidth}
		\centering
		\includegraphics[width=\linewidth]{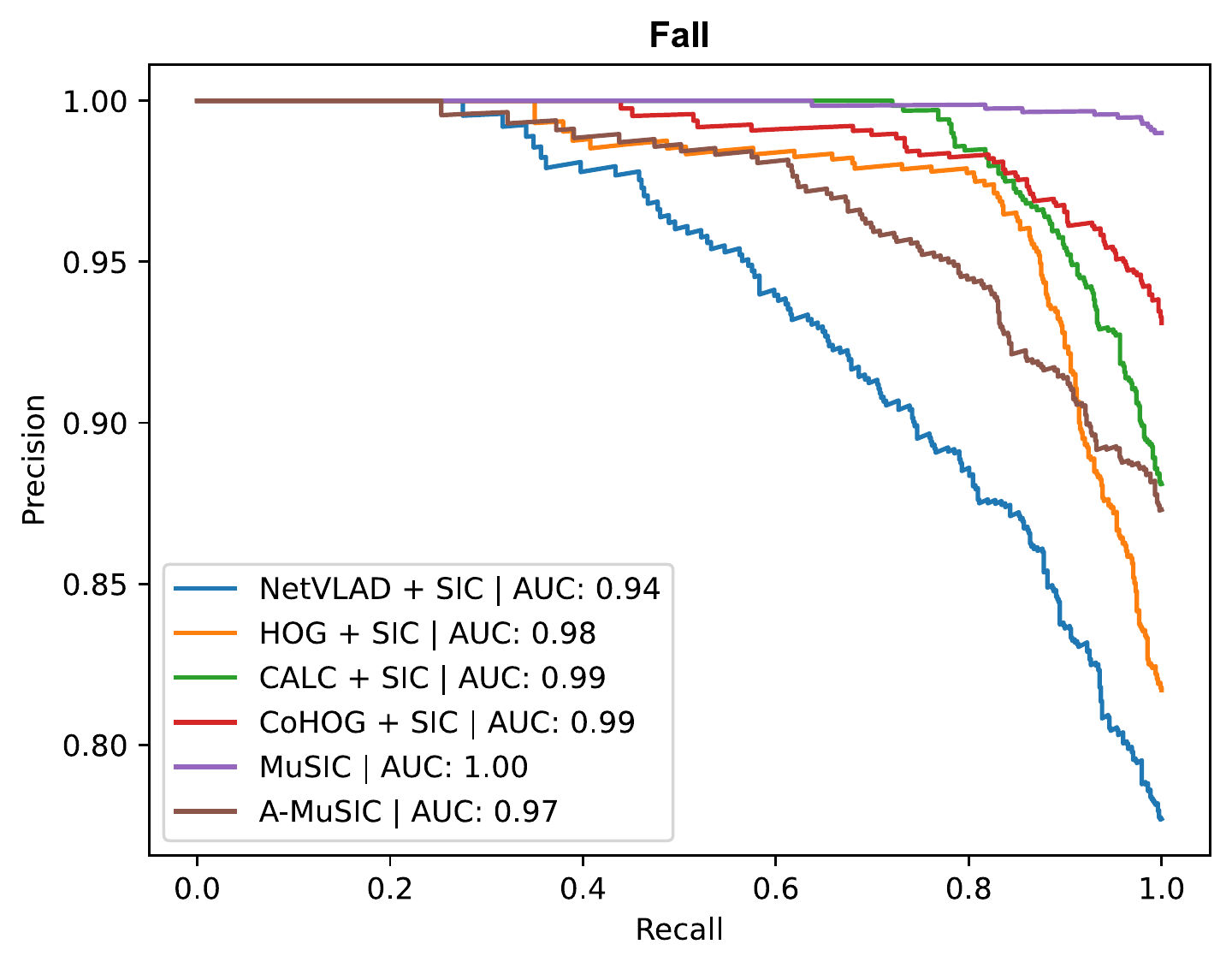}
		\caption{}
		\label{fig:pr_curves:B}
	\end{subfigure}
	\hfill	
	\begin{subfigure}[b]{0.32\textwidth}
		\centering
		\vspace{2ex}
		\includegraphics[width=\linewidth]{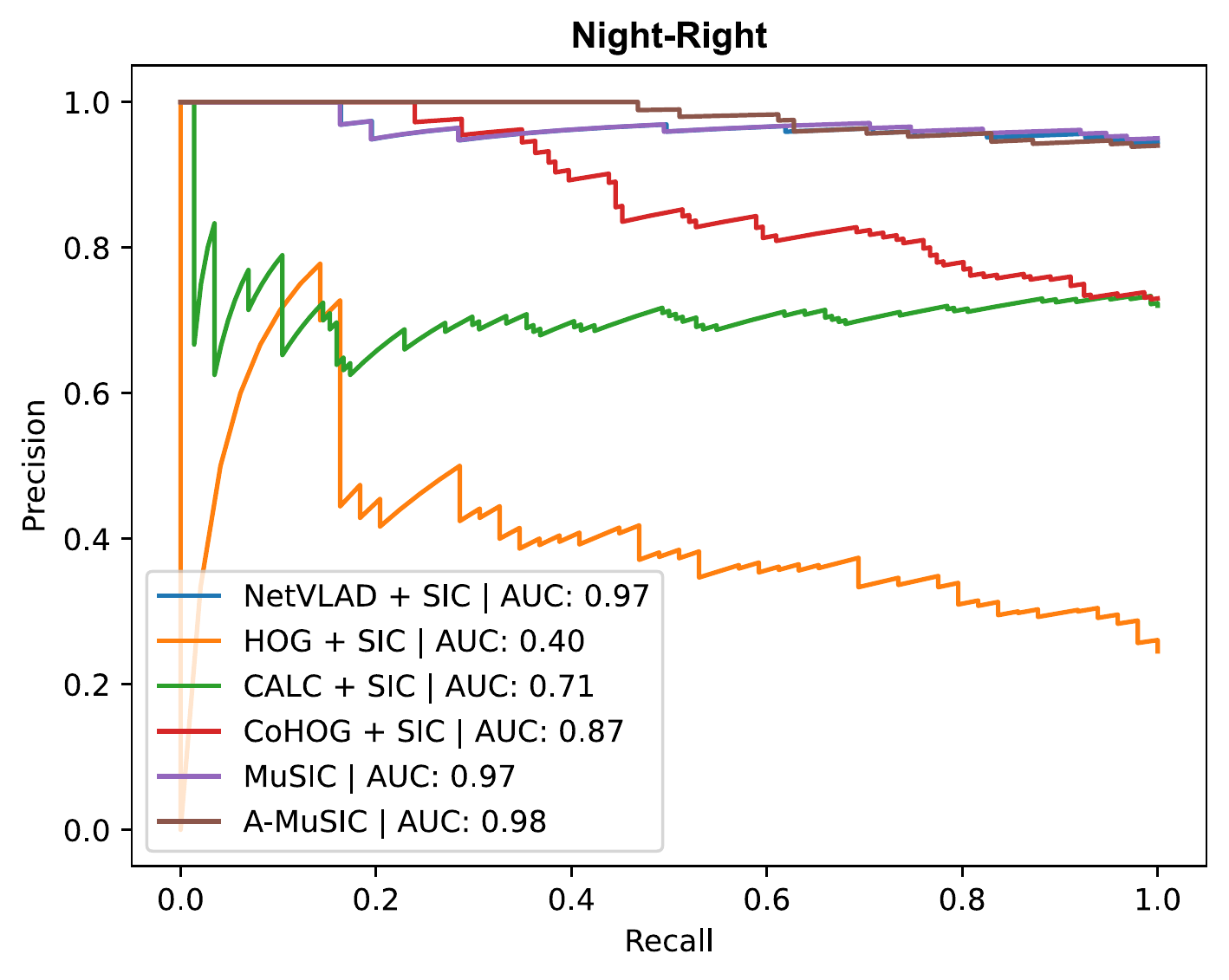}
		\caption{}
		\label{fig:pr_curves:C}
  
	\end{subfigure}	
        \begin{subfigure}[b]{0.32\textwidth}
		\centering
		\includegraphics[width=\linewidth]{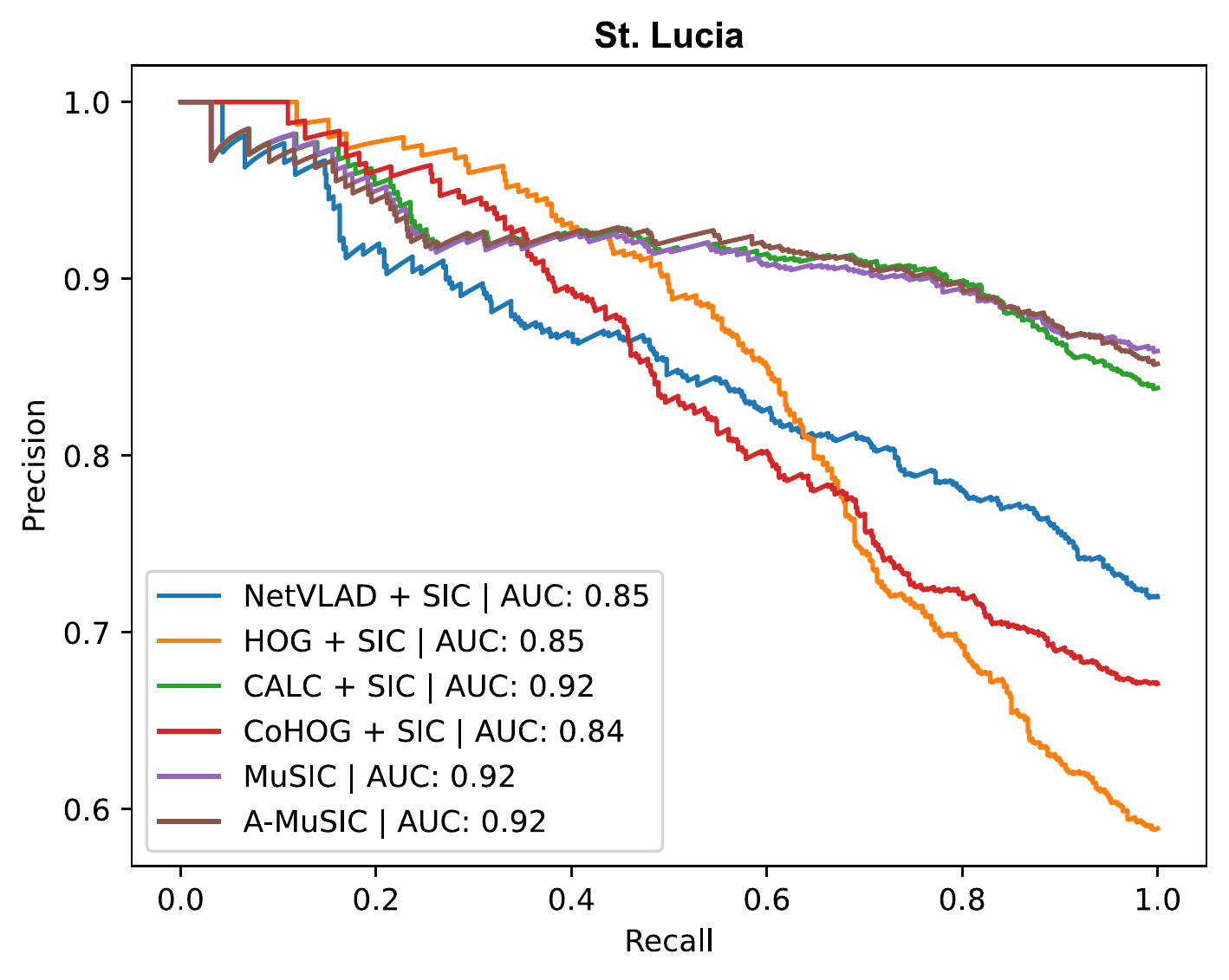}
		\caption{}
		\label{fig:pr_curves:D}
	\end{subfigure}
	\hfill
	\begin{subfigure}[b]{0.32\textwidth}
		\centering
		\includegraphics[width=\linewidth]{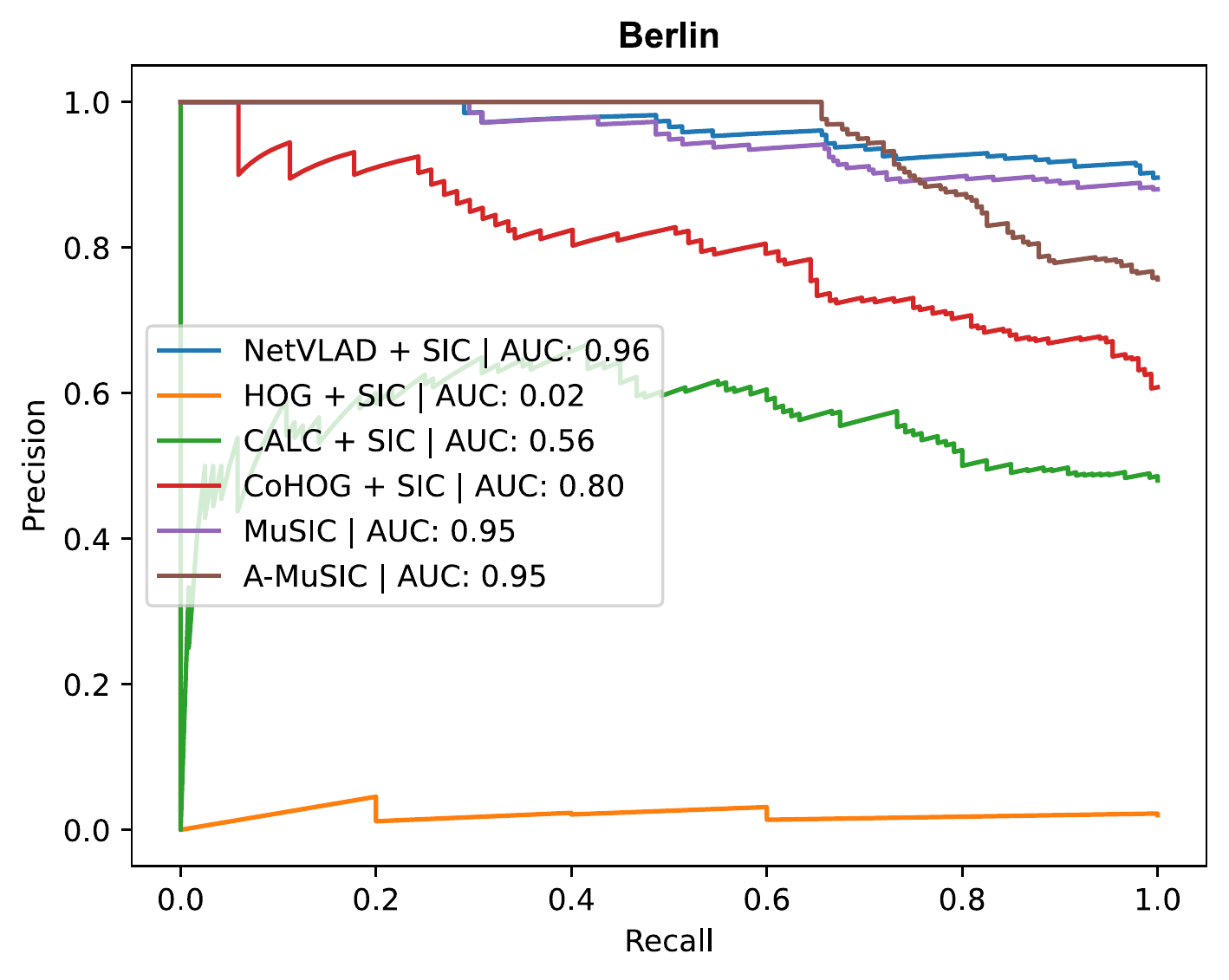}
		\caption{}
		\label{fig:pr_curves::E}
	\end{subfigure}
	\hfill	
	\begin{subfigure}[b]{0.32\textwidth}
		\centering
		\vspace{2ex}
		\includegraphics[width=\linewidth]{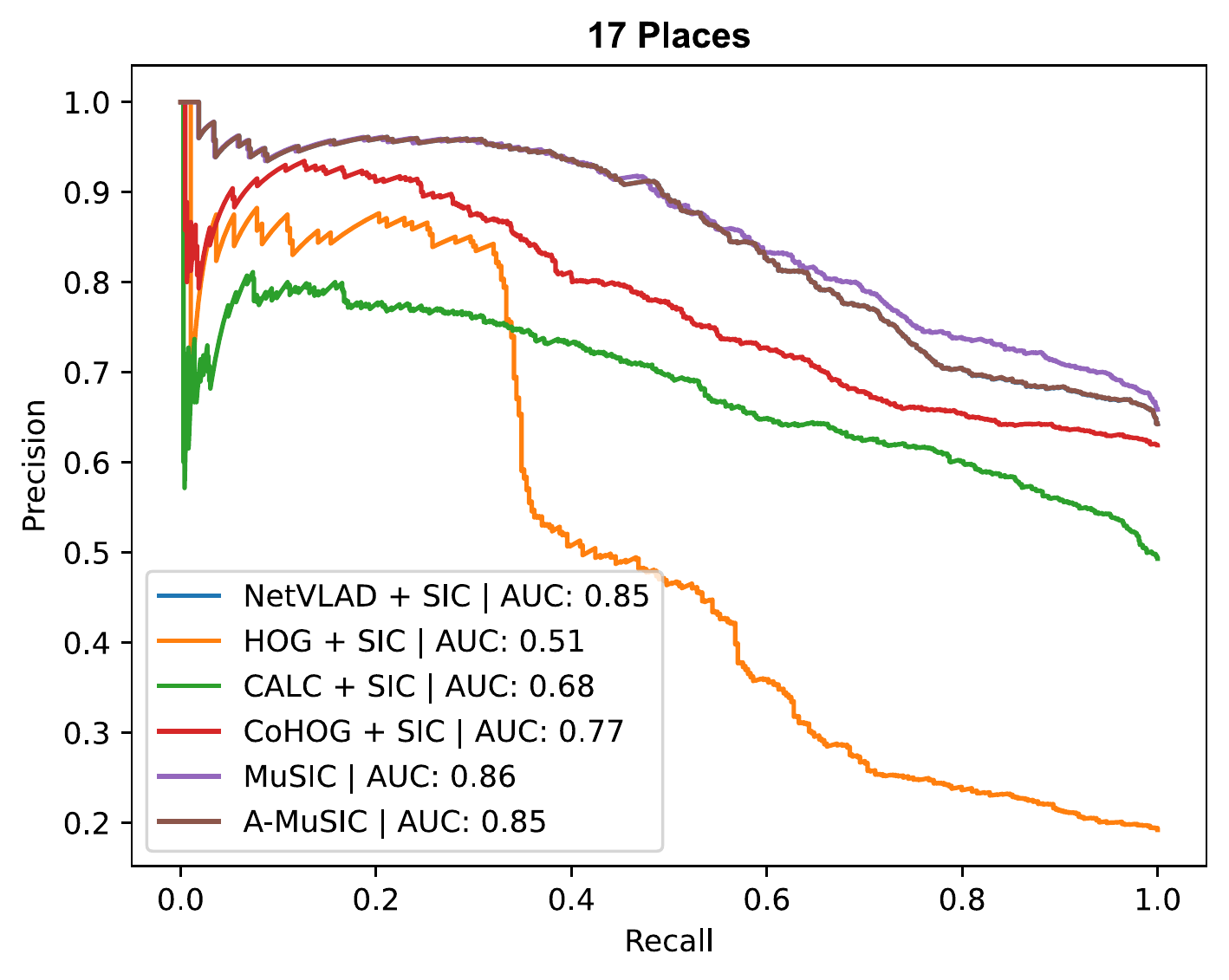}
		\caption{}
		\label{fig:pr_curves:F}
	\end{subfigure}	
	\caption{Precision-Recall Curves}
	\label{fig:pr_curves}
\end{figure*}

\subsection{MuSIC}
\begin{figure}[!t]
\vspace*{1ex}
\centering
\includegraphics[width=0.9\columnwidth]{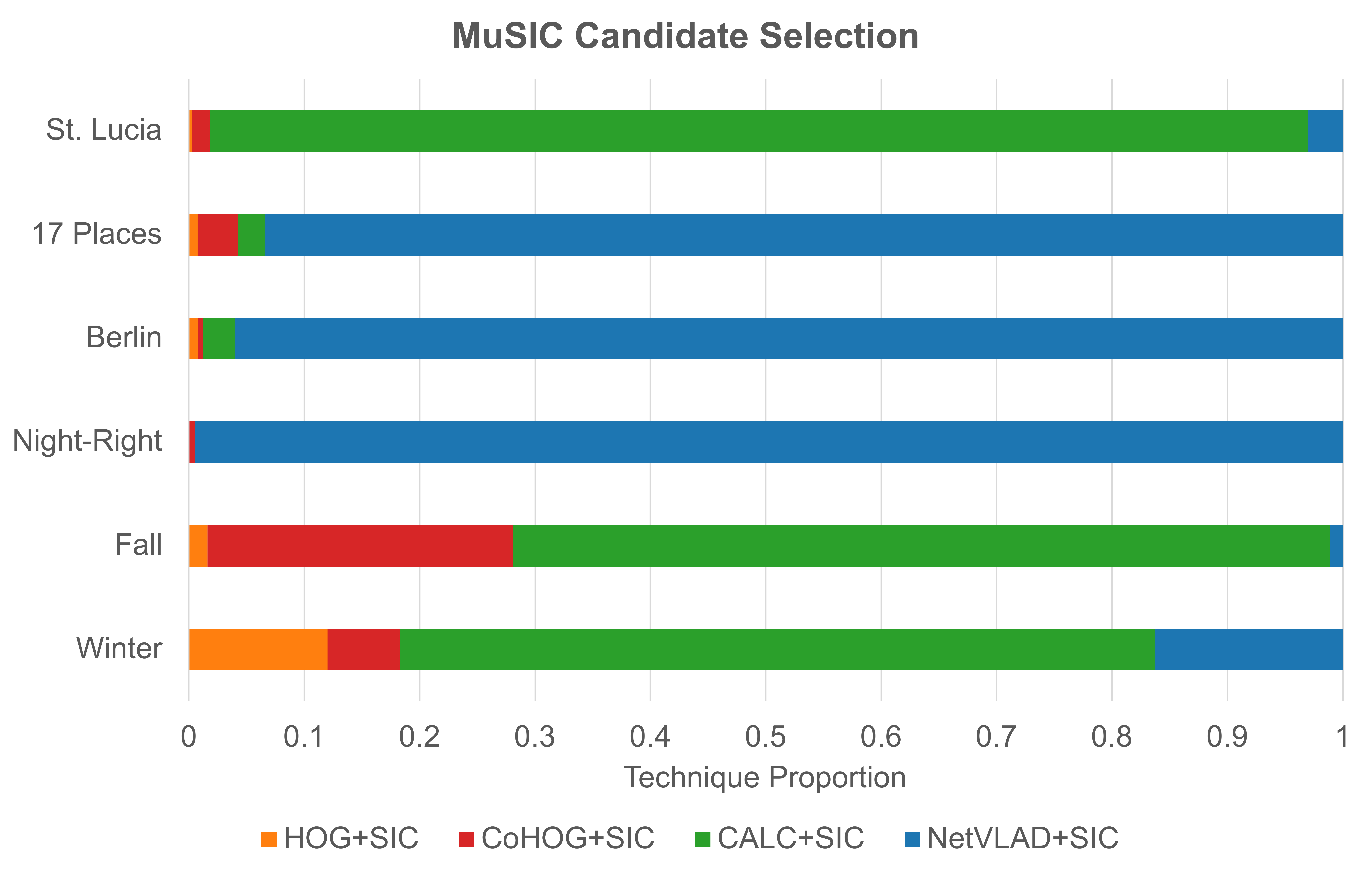}
\caption{MuSIC chosen candidate source technique proportion per dataset}
\label{fig:music_candidate_selection}
\end{figure}

MuSIC achieves stronger VPR performance across all datasets, observable in the PR curves of Fig. \ref{fig:pr_curves}. In Fig. \ref{fig:music_candidate_selection}, we can observe how each technique contributed to the increased performance of MuSIC. Note how in datasets where there is a clear dominant technique, MuSIC's PR-curve closely follows that technique's curve, and performance is similar. For example, in the 17 Places dataset, NetVLAD is almost exclusively chosen by MuSIC for every frame (Fig. \ref{fig:music_candidate_selection}) and their PR-curves are completely overlapped in Fig. \ref{fig:pr_curves:F}. On the other hand, when there is a larger variety of technique choice by MuSIC, there is a larger increase in VPR performance and its PR-curve does not follow any single technique. Such is the case of Winter, where all four techniques contribute to some extent, the performance increase is the most significant and MuSIC's curve quickly diverges from the remaining in Fig. \ref{fig:pr_curves:A}.

The increased VPR performance provided by MuSIC provides evidence that the method is able to choose the correct technique to place a particular match. However, MuSIC requires running all techniques for every query which results in extremely high prediction times. As expected, in table \ref{tab:vpr_perf_comp_times}, we observe that the prediction time of MuSIC is roughly the sum of times of the baseline techniques.

\subsection{A-MuSIC}
\begin{figure}[!t]
\vspace*{1ex}
\centering
\includegraphics[width=1.0\columnwidth]{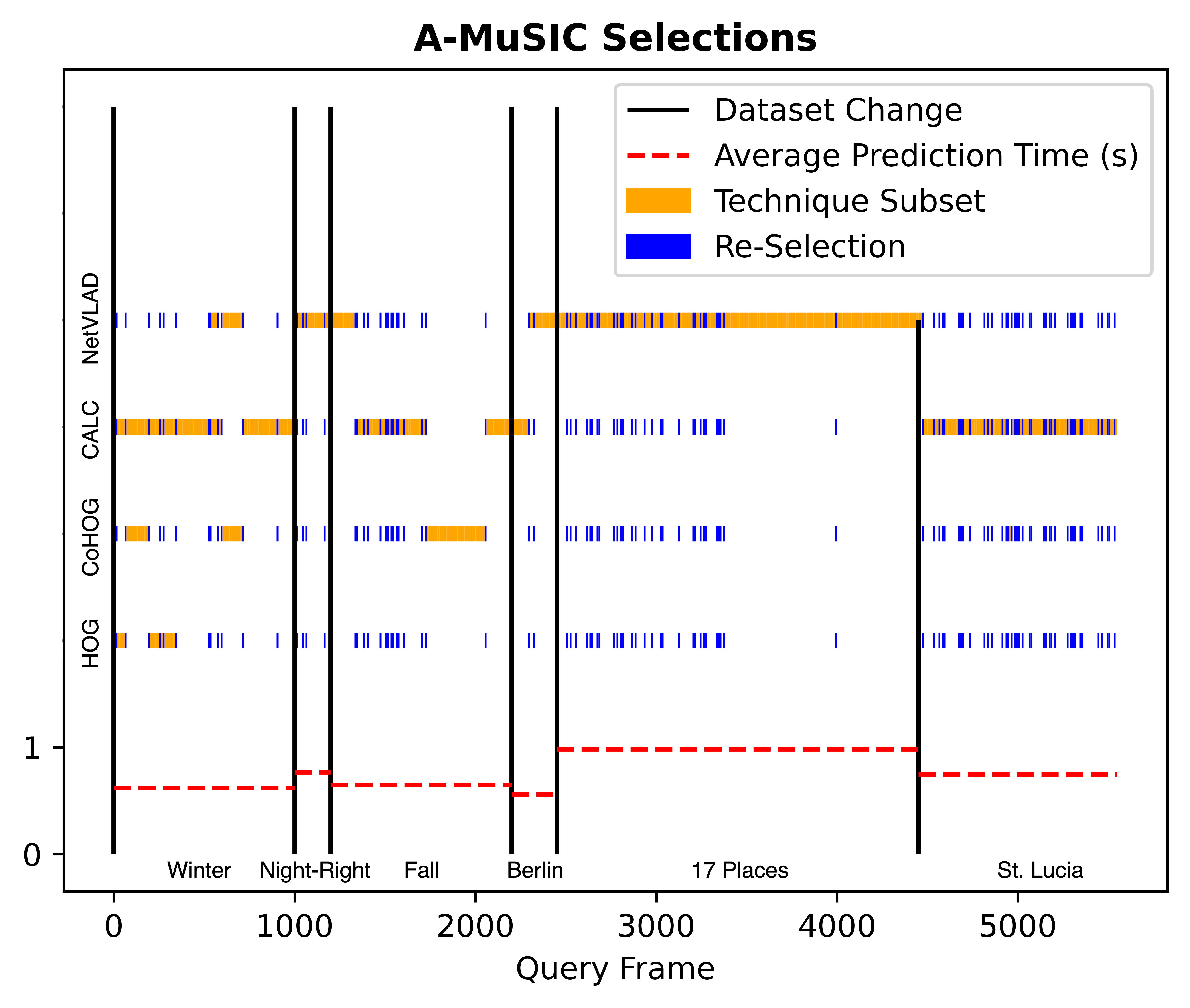}
\caption{A-MuSIC technique selections and re-selection timings}
\label{fig:amusic_selections}
\end{figure}

\newcolumntype{M}[1]{>{\centering\arraybackslash}m{#1}}
\begin{table}[t]
  \centering
  \caption{}
  \label{tab:PTR}%
    \begin{tabular}{M{1.5cm}M{1.5cm}M{1.5cm}M{1.5cm}}
    \toprule
Dataset     & PTR  & Technique Runs & Re-Selections\\
\midrule
Winter      & 0.45   & 1810      &   28      \\
Fall        & 0.36   &  1450       &  12    \\
Berlin      & 0.31    & 310          & 28  \\
Night-Right & 0.4     & 320             &  15 \\
St. Lucia   & 0.51     & 2240       &   235    \\
17 Places   & 0.36     &  2870      &    86   \\
Overall     & 0.41     & 9000         &   404   \\

\bottomrule
\end{tabular}
\end{table}%

In Table \ref{tab:vpr_perf_comp_times}, we observe that A-MuSIC achieves higher average VPR performance when compared to every SIC corrected technique. The same is true for performance on individual datasets with the exception of Fall. We can also see very similar average VPR performance between A-MuSIC and MuSIC, with a small drop of 0.01 AUC for the former. Looking at the Berlin, Night-Right, 17 Places and St. Lucia datasets, A-MuSIC achieves the same VPR performance as MuSIC, indicating that the sub-selection of techniques (Fig. \ref{fig:amusic_selections}) is working correctly. This is also visible in the corresponding PR curves in Fig. \ref{fig:pr_curves}, where the curve for A-MuSIC closely follows that of MuSIC. Winter is the dataset where the largest drop in performance occurs when compared to MuSIC, from an AUC value of 0.95 to 0.90, which is consistent with our findings in Fig. \ref{fig:music_candidate_selection}. Since Winter benefits the most from using multiple techniques, A-MuSIC's restriction on the amount of techniques used at any given time hurts its performance the most. Fall is the only dataset where A-MuSIC performs slightly worse than the best individual techniques, indicating that NetVLAD was erroneously selected for a portion of this dataset, visible in Fig. \ref{fig:amusic_selections} at the start of the Fall sequence.

by looking at the selection pattern of A-MuSIC, represented in Fig. \ref{fig:amusic_selections}, we can better understand the VPR performance of the adaptive method. Winter is the dataset where more techniques are selected in unison, with CALC still being on its own more often. On Night-Right, apart from re-selection periods, only NetVLAD is ran for the entire dataset. Fall's pattern shows that only one technique was selected at any point in time, but it alternated between NetVLAD, CoHOG and CALC. There was a short delay in the triggering of a re-selection when entering the Fall sequence, allowing for NetVLAD to perform VPR on its own for the beginning of the dataset and resulting in the small AUC drop reported. 17 Places clearly shows A-MUSIC's ability to use as little as a single technique for long periods, with only NetVLAD being used for the entire dataset. While St. Lucia displays more re-selection stages, only CALC was ever selected by the system to perform VPR.

The sub-selection of techniques described results in significant computation benefits, with the average prediction time of A-MuSIC being almost 1 second less than that of MuSIC (Table \ref{tab:vpr_perf_comp_times}). Moreover, the average prediction time of A-MuSIC is similar to that of NetVLAD, even being faster than the individual technique on some datasets. In Table \ref{tab:PTR}, we observe that the overall PTR of A-MuSIC is 0.41, effectively cutting $59\%$ of technique runs when compared to the non-adaptive MuSIC.

\section{Conclusions and Future Work}
\label{conclusions}
In this work, we propose a novel multi-technique VPR system capable of selecting a minimal optimal subset of techniques for the current environment without prior ground-truth knowledge. The size of the selection ranges from a single technique to the entire ensemble. Our approach analyses the sequential continuity of top candidates to identify incorrect matches at runtime, allowing a technique to assess its own online performance and attempt self-correction. Additionally, frame-by-frame analysis enables the correction method to utilize multiple techniques, improving overall VPR performance and gathering data on the most significant contributors in the current environment. The adaptive system examines both the correction information and contribution proportions, determining which techniques to run and identifying when a re-selection is necessary.

However, our adaptive system has clear limitations. It assumes sequential navigation, as the underlying correction method relies on analysing the sequential continuity of top predictions. The work could be improved by introducing a dynamic selection frame window for faster detection of changes in correction and triggering re-selection. Moreover, the quantification of correction should be refined for more detailed information on how techniques cope with the current environment, leading to more accurate re-selection timings.

This work demonstrates the advantages of an online-adaptive VPR system, increasing VPR performance by combining multiple techniques while minimizing unnecessary computation. 





\bibliographystyle{IEEEtran}
\typeout{}
\bibliography{ref}

\begin{thebibliography}{10}
\providecommand{\url}[1]{#1}
\csname url@rmstyle\endcsname
\providecommand{\newblock}{\relax}
\providecommand{\bibinfo}[2]{#2}
\providecommand\BIBentrySTDinterwordspacing{\spaceskip=0pt\relax}
\providecommand\BIBentryALTinterwordstretchfactor{4}
\providecommand\BIBentryALTinterwordspacing{\spaceskip=\fontdimen2\font plus
\BIBentryALTinterwordstretchfactor\fontdimen3\font minus
  \fontdimen4\font\relax}
\providecommand\BIBforeignlanguage[2]{{%
\expandafter\ifx\csname l@#1\endcsname\relax
\typeout{** WARNING: IEEEtran.bst: No hyphenation pattern has been}%
\typeout{** loaded for the language `#1'. Using the pattern for}%
\typeout{** the default language instead.}%
\else
\language=\csname l@#1\endcsname
\fi
#2}}

\bibitem{ref:vpr-survey}
S.~Lowry, N.~S{\"u}nderhauf, P.~Newman, J.~J. Leonard, D.~Cox, P.~Corke, and
  M.~J. Milford, ``Visual place recognition: A survey,'' \emph{IEEE
  Transactions on Robotics}, vol.~32, no.~1, pp. 1--19, 2015.

\bibitem{ref:illu_changes}
A.~Ranganathan, S.~Matsumoto, and D.~Ilstrup, ``Towards illumination invariance
  for visual localization,'' in \emph{2013 IEEE International Conference on
  Robotics and Automation}.\hskip 1em plus 0.5em minus 0.4em\relax IEEE, 2013,
  pp. 3791--3798.

\bibitem{ref:season_changes}
T.~Naseer, L.~Spinello, W.~Burgard, and C.~Stachniss, ``Robust visual robot
  localization across seasons using network flows,'' in \emph{Twenty-eighth
  AAAI conference on artificial intelligence}, 2014.

\bibitem{ref:pov_changes}
A.~Pronobis, B.~Caputo, P.~Jensfelt, and H.~I. Christensen, ``A discriminative
  approach to robust visual place recognition,'' in \emph{2006 IEEE/RSJ
  International Conference on Intelligent Robots and Systems}.\hskip 1em plus
  0.5em minus 0.4em\relax IEEE, 2006, pp. 3829--3836.

\bibitem{ref:maria_compl}
M.~Waheed, M.~J. Milford, K.~Mcdonald-Maier, and S.~Ehsan, ``Improving visual
  place recognition performance by maximising complementarity,'' \emph{IEEE
  Robotics and Automation Letters}, 2021\color{black}.

\bibitem{ref:mpf}
S.~Hausler, A.~Jacobson, and M.~Milford, ``Multi-process fusion: Visual place
  recognition using multiple image processing methods,'' \emph{IEEE Robotics
  and Automation Letters}, {2019}, volume=blue{4}, number=blue{2},
  pages={1924-1931}, doi={10.1109/LRA.2019.2898427 }.

\bibitem{hier_mpf}
S.~Hausler and M.~Milford, ``Hierarchical multi-process fusion for visual place
  recognition,'' in \emph{2020 IEEE International Conference on Robotics and
  Automation (ICRA)}, 2020, pp. 3327--3333.

\bibitem{switchhit}
M.~Waheed, M.~Milford, K.~McDonald-Maier, and S.~Ehsan, ``Switchhit: A
  probabilistic, complementarity-based switching system for improved visual
  place recognition in changing environments,'' in \emph{2022 IEEE/RSJ
  International Conference on Intelligent Robots and Systems (IROS)}, 2022, pp.
  7833--7840.

\bibitem{ref:fabmap}
M.~Cummins and P.~Newman, ``Appearance-only {SLAM} at large scale with
  {FAB-MAP} 2.0,'' \emph{The International Journal of Robotics Research},
  vol.~30, no.~9, pp. 1100--1123, 2011.

\bibitem{ref:sift}
D.~G. Lowe, ``Distinctive image features from scale-invariant keypoints,''
  \emph{International journal of computer vision}, vol.~60, no.~2, pp. 91--110,
  2004\color{black}.

\bibitem{ref:surf}
H.~Bay, A.~Ess, T.~Tuytelaars, and L.~Van~Gool, ``Speeded-up robust features
  {(SURF)},'' \emph{Computer vision and image understanding}, vol. 110, no.~3,
  pp. 346--359, 2008.

\bibitem{cummins2008fabmap}
M.~Cummins and P.~Newman, ``Fabmap: Probabilistic localization and mapping in
  the space of appearance,'' \emph{The International Journal of Robotics
  Research}, vol.~27, no.~6, pp. 647--665, 2008.

\bibitem{ref:hog}
N.~Dalal and B.~Triggs, ``Histograms of oriented gradients for human
  detection,'' in \emph{2005 IEEE computer society conference on computer
  vision and pattern recognition (CVPR'05)}, vol.~1.\hskip 1em plus 0.5em minus
  0.4em\relax Ieee, 2005, pp. 886--893.

\bibitem{ref:gist}
A.~Oliva and A.~Torralba, ``Building the gist of a scene: The role of global
  image features in recognition,'' \emph{Progress in brain research}, vol. 155,
  pp. 23--36, 2006.

\bibitem{ref:mcmanus2014scene}
C.~McManus, B.~Upcroft, and P.~Newmann, ``Scene signatures: Localised and
  point-less features for localisation,'' in \emph{Proceedings of Robotics:
  Science and Systems}, Berkeley, USA, July 2014.

\bibitem{ref:cohog}
M.~Zaffar, S.~Ehsan, M.~Milford, and K.~McDonald-Maier, ``{CoHOG}: A
  light-weight, compute-efficient, and training-free visual place recognition
  technique for changing environments,'' \emph{IEEE Robotics and Automation
  Letters}, vol.~5, no.~2, pp. 1835--1842, 2020.

\bibitem{rmac}
G.~Tolias, R.~Sicre, and H.~J{\'e}gou, ``Particular object retrieval with
  integral max-pooling of cnn activations,'' in \emph{ICLR 2016-International
  Conference on Learning Representations}, 2016, pp. 1--12.

\bibitem{ref:cnn_for_vpr1}
Y.~{Hou}, H.~{Zhang}, and S.~{Zhou}, ``Convolutional neural network-based image
  representation for visual loop closure detection,'' in \emph{2015 IEEE
  International Conference on Information and Automation}, 2015, pp.
  2238--2245.

\bibitem{ref:cnn_for_vpr2}
D.~Bai, C.~Wang, B.~Zhang, X.~Yi, and X.~Yang, ``Sequence searching with cnn
  features for robust and fast visual place recognition,'' \emph{Computers \&
  Graphics}, vol.~70, pp. 270--280, 2018.

\bibitem{ref:hybridasmosnet}
Z.~Chen, A.~Jacobson, N.~S{\"u}nderhauf, B.~Upcroft, L.~Liu, C.~Shen, I.~Reid,
  and M.~Milford, ``Deep learning features at scale for visual place
  recognition,'' in \emph{2017 IEEE International Conference on Robotics and
  Automation (ICRA)}.\hskip 1em plus 0.5em minus 0.4em\relax IEEE, 2017, pp.
  3223--3230.

\bibitem{ref:netvlad}
R.~Arandjelovic, P.~Gronat, A.~Torii, T.~Pajdla, and J.~Sivic, ``{NetVLAD}:
  {CNN} architecture for weakly supervised place recognition,'' in
  \emph{Proceedings of the IEEE conference on computer vision and pattern
  recognition}, 2016, pp. 5297--5307.

\bibitem{ref:calc}
N.~Merrill and G.~Huang, ``Lightweight unsupervised deep loop closure,''
  \emph{arXiv preprint arXiv:1805.07703}, 2018.

\bibitem{ref:vpr_uavs}
B.~Ferrarini, M.~Waheed, S.~Waheed, S.~Ehsan, M.~Milford, and K.~D.
  McDonald-Maier, ``Visual place recognition for aerial robotics: Exploring
  accuracy-computation trade-off for local image descriptors,'' in \emph{2019
  NASA/ESA Conference on Adaptive Hardware and Systems (AHS)}.\hskip 1em plus
  0.5em minus 0.4em\relax IEEE, 2019, pp. 103--108.

\bibitem{ref:nord}
N.~S{\"u}nderhauf, P.~Neubert, and P.~Protzel, ``Are we there yet? challenging
  {SeqSLAM} on a 3000 km journey across all four seasons,'' in \emph{Proc. of
  Workshop on Long-Term Autonomy, IEEE International Conference on Robotics and
  Automation (ICRA)}.\hskip 1em plus 0.5em minus 0.4em\relax Citeseer, 2013, p.
  2013.

\bibitem{stlucia}
A.~J. Glover, W.~P. Maddern, M.~J. Milford, and G.~F. Wyeth, ``Fab-map+
  ratslam: Appearance-based slam for multiple times of day,'' in \emph{2010
  IEEE international conference on robotics and automation}.\hskip 1em plus
  0.5em minus 0.4em\relax IEEE, 2010, pp. 3507--3512.

\bibitem{17places}
R.~Sahdev and J.~K. Tsotsos, ``Indoor place recognition system for localization
  of mobile robots,'' in \emph{2016 13th Conference on computer and robot
  vision (CRV)}.\hskip 1em plus 0.5em minus 0.4em\relax IEEE, 2016, pp. 53--60.

\bibitem{berlin}
Z.~Chen, F.~Maffra, I.~Sa, and M.~Chli, ``Only look once, mining distinctive
  landmarks from convnet for visual place recognition,'' in \emph{2017 IEEE/RSJ
  International Conference on Intelligent Robots and Systems (IROS)}.\hskip 1em
  plus 0.5em minus 0.4em\relax IEEE, 2017, pp. 9--16.

\bibitem{ref:pr_jus1}
J.~Davis and M.~Goadrich, ``The relationship between {Precision-Recall} and
  {ROC} curves,'' in \emph{Proceedings of the 23rd international conference on
  Machine learning}, 2006, pp. 233--240.

\bibitem{ref:vpr_bench}
M.~Zaffar, S.~Garg, M.~Milford, J.~Kooij, D.~Flynn, K.~McDonald-Maier, and
  S.~Ehsan, ``Vpr-bench: An open-source visual place recognition evaluation
  framework with quantifiable viewpoint and appearance change,''
  \emph{International Journal of Computer Vision}, pp. 1--39, 2021.

\end{thebibliography}

\end{document}